%% file: neurips_2020.tex
\documentclass{article}
\usepackage[preprint, nonatbib]{neurips_2020}

\usepackage[utf8]{inputenc} 
\usepackage[T1]{fontenc}    
\usepackage{url}            
\usepackage{booktabs}       
\usepackage{amsfonts}       
\usepackage{nicefrac}       
\usepackage{microtype}      
\usepackage{amsmath}
\usepackage{caption}
\usepackage{multirow}
\usepackage{subcaption}
\usepackage[dvipsnames]{xcolor}
\usepackage{xspace}
\usepackage{hyperref}
\hypersetup{pagebackref=true,breaklinks=true,letterpaper=true,colorlinks,bookmarks=false}
\usepackage{amssymb}

\usepackage[citestyle=numeric,maxbibnames=9,maxcitenames=2,natbib=true,backend=bibtex,backref=true]{biblatex}
\usepackage{subcaption}
\usepackage{graphicx}

\newcommand{\mbf}[1]{\mathbf{#1}}

\newcommand{\h}{\mathbf{H}}
\newcommand{\htt}{\mathbf{H_t}}
\newcommand{\xtt}{\mathbf{X_t}}
\newcommand{\ctt}{\mathbf{c_t}}
\newcommand{\sigmoid}{\mathbf{\sigma}}
\newcommand{\ml}{MarkedLong\xspace}
\newcommand{\pf}{PathFinder\xspace}

\newcommand{\ig}{\mathbf{i}}
\newcommand{\fg}{\mathbf{f}}
\newcommand{\og}{\mathbf{o}}
\newcommand{\ggt}{\mathbf{g}}
\newcommand{\wxh}{\mathbf{W_{xh}}}
\newcommand{\uhh}{\mathbf{U_{hh}}}

\newcommand{\cell}{\mathbf{c}}

\newcommand{\subt}{\mathbf{t}}

\newcommand{\het}{\mathbf{H_t^{exc}}}
\newcommand{\hit}{\mathbf{H_t^{inh}}}
\newcommand{\hdt}{\mathbf{H_t^{div}}}

\newcommand{\wexc}{\mathbf{W_{exc}}}
\newcommand{\winh}{\mathbf{W_{inh}}}
\newcommand{\wdiv}{\mathbf{W_{div}}}

\newcommand{\R}{\mathbb{R}}

\definecolor{exc_color}{rgb}{1.0, 0.0, 0.0}
\definecolor{inh_color}{rgb}{0.2, 0.8, 0.4}
\definecolor{div_color}{rgb}{0.2, 1.0, 0.4}

\title{Learning compact generalizable neural representations supporting perceptual grouping}

\author{%
  Vijay Veerabadran$^{*}$, 
  Virginia R. de Sa$^{*+}$ \\
  $^*$Department of Cognitive Science, $^+$
  Hal\i c\i o\u{g}lu Data Science Institute, \\
  University of California, San Diego\\
  La Jolla, CA 92093 \\
  \texttt{vveeraba@ucsd.edu}
  }

\begin{document}

\maketitle

\begin{abstract}
Work at the intersection of vision science and deep learning is starting to explore the efficacy of deep convolutional networks (DCNs) and recurrent networks in solving perceptual grouping problems that underlie primate visual recognition and segmentation \cite{Brosch2015ReinforcementNetworks,linsley2018learning}. Here, we extend this line of work to investigate the compactness and generalizability of DCN solutions to learning low-level perceptual grouping routines involving contour integration. We introduce V1Net, a bio-inspired recurrent unit that incorporates lateral connections ubiquitous in cortical circuitry. 
Feedforward convolutional layers in DCNs can be substituted with V1Net modules to enhance their contextual visual processing support for perceptual grouping. 
We compare the learning efficiency and accuracy of V1Net-DCNs to that of 14 carefully selected feedforward and recurrent neural architectures (including state-of-the-art DCNs) on MarkedLong -- a synthetic forced-choice contour integration dataset of 800,000 images we introduce here -- and the previously published Pathfinder contour integration benchmarks \cite{linsley2018learning}. 
We gauged solution generalizability by measuring the transfer learning performance of our candidate models trained on MarkedLong that were fine-tuned to learn PathFinder. Our results demonstrate that a compact 3-layer V1Net-DCN matches or outperforms the test accuracy and sample efficiency of all tested comparison models which contain between 5x and 1000x more trainable parameters; we also note that V1Net-DCN learns the most compact generalizable solution to \ml. 
A visualization of  the temporal dynamics of a V1Net-DCN  
elucidates its usage of interpretable grouping computations to solve \ml. 
The compact and rich representations of V1Net-DCN also make it a promising candidate to build on-device machine vision algorithms as well as help better understand biological cortical circuitry.
\end{abstract}
\input{introduction}
\input{background}
\input{methods_v1net}
\input{methods_markedlong}
\input{experiments}
\input{discussion}
\input{acknowledgments}
\section*{Broader Impact}
This work has the potential for broader impacts in four different areas.
First, studying these networks and their strengths and deficiencies may reveal new insights into biological cortical processing. However these insights should be done carefully without making unfounded claims.
Second, V1Net by virtue of its small number of parameters may provide powerful performance  on memory-limited edge devices.
Third, the V1Net network, or a future improved version, may improve deep learning performance and transferability on tasks that involve contour and contextual integration.   
The emphasis on contour and shape processing, may reduce the "texture bias" seen in feedforward deep networks and may even, by virtue of using a more human-like architecture, reduce the susceptibility to adversarial examples which emphasize differences between computer and human recognition.  Again this hypothesis should be carefully investigated and tested before implemented on any critical software.
Fourth, the introduction of the MarkedLong dataset could be useful for better exploring and understanding the processing capabilities and processing in deep convolutional networks.

\printbibliography
\input{neurips_2020_si}
\end{document}

%% file: introduction.tex
\section{Introduction}
\label{sec:intro}
Following Hubel and Wiesel's \citep{hubel1968receptive} seminal work on characterizing receptive fields in the cat striate cortex and Fukushima's Neocognitron \citep{fukushima1980neocognitron} (a hierarchical extension of this building block), two broad families of visual models have evolved from the neuroscience and computer vision communities respectively. The former family of models aims to account for findings from neurophysiology, either by directly modeling types of neuronal responses \citep{de1982spatial,sullivan2006model,chichilnisky2001simple,pillow2007likelihood} or by proposing computational models that explain perceptual phenomena \citep{blakeslee1999multiscale,robinson2007}.

The latter family of computer vision models, particularly Deep Convolutional Networks (DCNs) \cite{lecun1998gradient}, 
are aimed at optimizing performance on various computer vision benchmarks including but not limited to image recognition \citep{krizhevsky2012imagenet,hu2018squeeze,Xie2019Self-trainingClassification,}, contour detection \citep{xie2015holistically,Shen2015DeepContourAD} and object segmentation \citep{He2020MaskR-CNN,long2015fully}. Various specialized architectural innovations \cite{He2020MaskR-CNN,Chollet2017Xception:Convolutions,Xie2017AggregatedRT,Elsayed2020RevisitingConnectivity}  integrated into DCNs --
a simplistic stack of convolutional operations, nonlinearities, divisive normalization (such as \cite{szegedy2015going, ba2016layer, Wu2018GroupNormalization}) and spatial pooling \cite{Zhang2019MakingAgain} combined with gradient-based optimization algorithms \cite{Ruder2016AnAlgorithms} --
 are bringing them closer to matching and at times outperforming humans on visual tasks where primate visual cortex excels. On the other hand, recent work highlights how DCNs also 
respond differently to form-preserving manipulations of natural images
 relative to biological vision \citep{geirhos2018imagenettrained,Serre2019DeepUgly,Geirhos2018GeneralisationNetworks,Xu2020LimitedNetworks,Baker2018DeepShape}. 
 These differences raise concerns about the extent to which
 processes that support visual recognition overlap mechanistically in DCNs and humans. 
 Do DCNs evolve computational strategies that implicitly utilize functionalities that are cortically implemented by nuanced biological visual processes such as recurrence, attentional modulation, and episodic memory? 
Recent work at the intersection of deep learning and vision science has started studying the recurrence aspect of this question by analyzing feedforward and recurrent DCN solutions to perceptual grouping problems \cite{Kim2019DisentanglingGrouping,Funke2020ThePerception} (i.e, problems on grouping parts of objects together into their respective wholes that are reportedly supported by recurrent cortical computations, see \cite{ullman1987visual,Roelfsema2011IncrementalVision,Kreiman2020BeyondCortex,Lamme1998FeedforwardCortex}). 
Here, we advance this research by making the following three-fold contribution:  (1) We introduce V1Net -- a novel bio-inspired recurrent unit based on prior research concerning the role of recurrent horizontal connections in human perceptual grouping, (2) We introduce \ml, a simple contour integration 
task formulated as a binary image classification problem; this proposed dataset contains 800,000 distinct stimulus images with a resolution of 256 $\times$ 256 pixels, (3) We report observations from our comprehensive comparison of V1Net's performance accuracy, sample efficiency, solution compactness (defined by number of trainable parameters) and solution generalization against that of a range of feedforward and recurrent neural architectures on \ml and the previously published \pf challenges \cite{linsley2018learning}. \\
Our results suggest that (i) V1Net, a carefully designed recurrent unit inspired by cortical long-range horizontal connectivity matches or outperforms task performance of all our comparison models while containing a fraction of their parameter count, (ii) V1Net's superior performance compared to two relatively parameter-heavy recurrent architectures highlights the importance of explicit linear-nonlinear horizontal interactions and gain modulation, and (iii) V1Net's horizontal connections are reminiscent of reports on their structure from single-cell physiology and psychophysics \cite{Gilbert1989ColumnarCortex,Field1993ContourField}.

%% file: background.tex
\section{Background}
Neuroscience and vision scientists have used recurrent computer vision models operating on static images 
to model biological horizontal connections and/or  re-entrant feedback connections. Typically in such models, bottom-up representations of static images are processed by a stack of identical computational blocks with weight sharing across the stack to model a nonlinear dynamical function of the bottom-up input (e.g., \cite{Liao2016BridgingCortex,perona1990scale}).
This approach is in contrast to treating static images as a temporal sequence of successive pixels \cite{VanDenOord2016PixelNetworks,Gregor2015DRAW:Generation,Oord2016ConditionalDecoders}.
\citet{zamir2017feedback} developed  a variant of the Convolutional-LSTM architecture to imbibe the intrinsic compositionality present in object features into recurrent neural representations through top-down feedback connections in their recurrent cell. 
CORNet-S proposed by \citet{Kubilius2018CORnet:Recognition} is a recurrent-convolutional model built through convolutional layers with residual skip-connections that
tops the Brain-Score leaderboard, a benchmark evaluating object recognition performance and match to physiological recordings along the  ventral visual pathway. 
\citet{tdcrnn} proposed a class of recurrent computer vision models similar to CORNet-S  designed through a large-scale Neural Architecture Search to jointly optimize object recognition performance and cortical resemblance.
Unlike the above-mentioned models, our proposed V1Net block explicitly incorporates linear-nonlinear inhibitory and excitatory horizontal connections along with a gain control mechanism; we empirically show that this implementation-level bio-inspired design leads to fast learning of grouping routines.
Horizontal Gated Recurrent Unit (hGRU), a model of cortical long-range horizontal connections proposed by \citet{linsley2018learning}  efficiently performs low-level perceptual grouping with   high accuracy and sample efficiency
compared to a range of feedforward and recurrent comparison models 
on their \pf challenge. 
Our proposed V1Net architecture is a computationally simpler and parameter-efficient model of horizontal connections compared to hGRU. 
Unlike one of hGRU's key assumptions that horizontal connections are symmetric in nature, V1Net incorporates 3 diverse cell types (and convolution kernels) which are not constrained to maintain symmetric horizontal synapses. Each of these cell types support one of three kinds of linear and nonlinear horizontal connectivity patterns. 
We explore key machine learning innovations that are lacking in hGRU. Among such innovations, V1Net utilizes depthwise-separable convolutions that have been shown to maintain high expressivity with a significant reduction in parameter count \cite{Chollet2017Xception:Convolutions, Howard2017MobileNets:Applications}. hGRU batch-normalizes the recurrent output state at each timestep using an independent batch normalization layer, restricting its inference operation to a fixed number of recurrent iterations that was used during training. V1Net tackles this issue by employing a single Layer Normalization operation \cite{ba2016layer} shared across timesteps.

\citet{Hasani2019SurroundNetworks} develop Surround Modulation Networks which introduce  non-trainable filter banks of feedforward and lateral connections ( similar to those in \citet{robinson2007}) to DCNs, resulting in improved sample efficiency and perceptual robustness on ILSVRC \cite{ILSVRC15}.
Feedforward Atrous convolution operations are another way to encode long-range spatial dependencies as has been reported in 
dense prediction and autoregressive modeling studies \cite{Yu2015Multi-ScaleConvolutions,Chen2018DeepLab:CRFs,Chen2018Encoder-decoderSegmentation,Oord2016WaveNet:Audio}. 
Conditional random fields are a class of probabilistic graphical models which are also of relevance and have been shown to improve performance on dense prediction problems such as semantic segmentation through incorporation of global visual context \cite{Arnab2018ConditionalPrediction, triggs2008scene}. 

%% file: methods_v1net.tex
\section{Designing a recurrent architecture implementing horizontal interactions}
\label{sec:v1net}
V1Net, our novel recurrent unit is designed to
merge the following three explicit bio-inspired functional modeling constraints with the convolutional variant of an LSTM cell (ConvLSTM) \cite{xingjian2015convolutional}: (a) Modeling of the push-pull interaction between cortical excitatory and inhibitory cell populations, (b) Modeling of three different cell types, each of which implement a particular kind of horizontal computation, (c) Incorporating a learned gain-control mechanism to stabilize horizontal population activity. Our horizontal connections learn to implement a version of incremental grouping by activity enhancement as proposed by \citet{Roelfsema2006CORTICALGROUPING}. Our experiments show that V1Net learns to build long-range horizontal interactions wherein spatially distant neurons that are part of the same object mutually excite each other along a transitive chain of local connections. This enhanced activity of neurons responding to parts of the same whole representationally implements a task-relevant grouping of bottom-up features;
we discuss this observation in Fig. \ref{fig:v1net_activations} and  Sec. \ref{sec:grouping_vis}.

\textbf{Mathematical formulation: }
V1Net's gating computation (Eqn. \ref{eqn:gating}) is identical to that of a ConvLSTM whose state updates are governed by the original LSTM update equations \cite{hochreiter1997long} with fully-connected operations replaced with convolution operations. At any discrete recurrent iteration $\subt$, consider the neuronal population activity pre-horizontal modulation to be available in the recurrent input $\xtt \in \R_{n,k,h,w}$, which is typically the output of a convolution layer in a DCN.
Contextual interactions between the $k^2$ input feature pairs in $\xtt$ are governed by V1Net's horizontal connections.
\begin{figure}
    \centering
        \includegraphics[width=\linewidth]{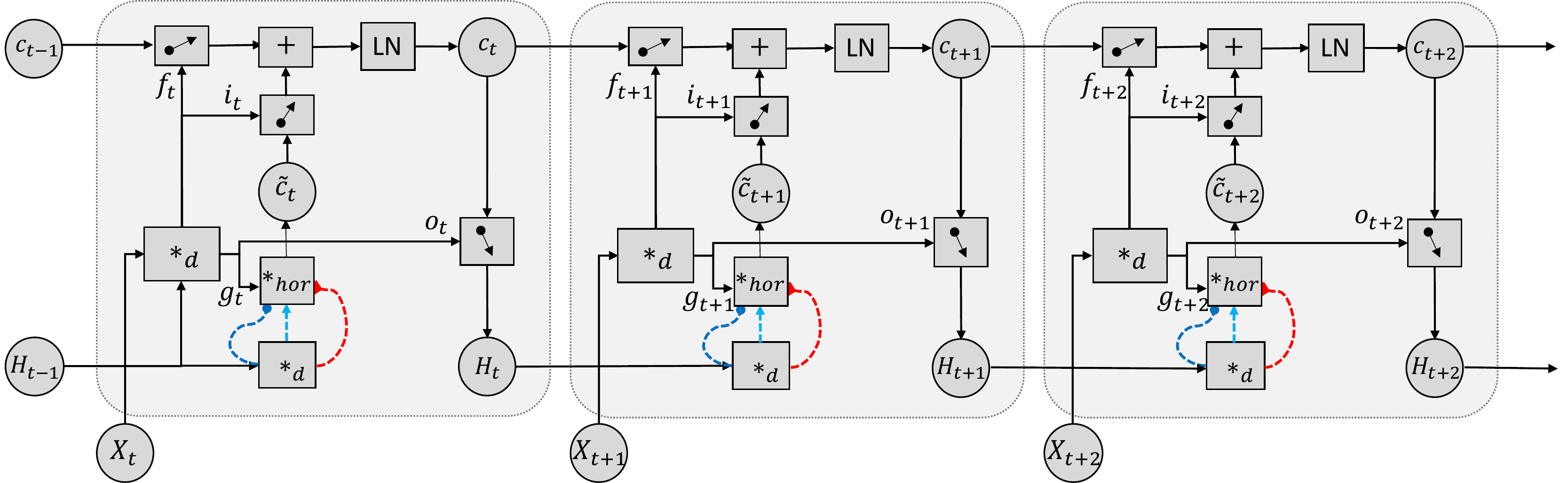}
        \caption{Our proposed V1Net cell architecture with recurrent linear excitatory (red), subtractive inhibitory (turquoise) and nonlinear gain modulating (navy blue) horizontal connections unrolled for three recurrent iterations. Eqn. \ref{eqn:linear_hor},\ref{eqn:gain_control},\ref{eqn:grouping_curr} from Sec. \ref{sec:v1net} are encapsulated into box ($*_{hor}$) in this figure.} 
        \label{fig:v1net_arch}
\end{figure} \\
The following procedure summarizes how V1Net's grouping representation is formed/updated at every recurrent iteration: The most recent neuronal response reflecting iterative grouping until iteration $\subt$ is available in V1Net's hidden state, $\htt$. Distinct convolutional filter banks with varying spatial dimensions, namely $\wdiv$, $\winh$ and $\wexc$, model a mapping from state $\h_{\mbf{\subt-1}}$ (from the previous iteration) and the bottom-up feedforward input $\xtt$ to the following populations of three different cell types each of which implement (1) gain control, (2) subtractive inhibition and (3) additive excitation. Their corresponding activities are stored in $\hdt$, $\hit$ and $\het$ (Eqn. \ref{eqn:linear_hor},\ref{eqn:gain_control}); each $\h^+_\subt$ is a 4-dimensional tensor of identical shape to $\htt$ and $\xtt$. Excitatory influence and gain modulation are then integrated with $\ggt_\subt$, a nonlinear function of $\xtt$ and $\htt$ to produce $\tilde{\cell}_\subt$, a candidate representation of the current grouping calculation (Eqn. \ref{eqn:grouping_curr}). $\tilde{\cell}_\subt$ is mixed with the previous grouping estimate ($\mbf{c_{t-1}}$) to produce an updated grouping estimate, $\ctt$. A nonlinear mapping of $\ctt$ following layer normalization (LN) and gating produces $\htt$ (Eqn. \ref{eqn:mixing}) which contains updated grouping information 
for subsequent processing. The following equations summarize V1Net's horizontal connection dynamics. Depthwise separable convolutions are accompanied by additive biases which are omitted from the equations for reading convenience:

\begin{equation}
\label{eqn:gating}
\begin{pmatrix}
\fg_\subt\\ 
\ig_\subt\\ 
\og_\subt\\ 
\ggt_\subt
\end{pmatrix} = \sigmoid(\wxh *_d \xtt + \uhh *_d \mbf{H_{t-1}}) 
\end{equation}
\begin{equation}
\label{eqn:linear_hor}
\textcolor{red}{\het}, \textcolor{cyan}{\hit} = \sigmoid(\wexc *_d \mbf{H_{t-1}}), \sigmoid(\winh *_d \mbf{H_{t-1}})
\end{equation}
\begin{equation}
\label{eqn:gain_control}
\textcolor{RoyalBlue}{\hdt} =\sigmoid(\wdiv *_d \mbf{H_{t-1}})
\end{equation}
\begin{equation}
\label{eqn:grouping_curr}
\tilde{\cell}_\subt = \textcolor{RoyalBlue}{\hdt} \times (\ggt_\subt + \textcolor{red}{\het}) - \textcolor{cyan}{\hit}; 
\end{equation}
\begin{equation}
\label{eqn:mixing}
\ctt = \fg_\subt\odot \cell_{t-1} + \ig_\subt \odot \tanh(\tilde{\cell}_\subt); \htt = \og_\subt \odot \mbf{\gamma}(\mbf{LN(\ctt)})
\end{equation}
Each $\mbf{W}_.$ and $\mbf{U}_.$ is a 2-D convolution kernel, $\sigmoid(.)$ and $\gamma(.)$ 
represent sigmoid and ReLU nonlinearities respectively; $*_d$ represents 2-D depthwise separable convolution \cite{Jin2014FlattenedAcceleration,Chollet2017Xception:Convolutions}. 
Color coding in Eqn. \ref{eqn:linear_hor},\ref{eqn:gain_control},\ref{eqn:grouping_curr} are consistent with that of Fig. \ref{fig:v1net_arch}.
A key property  is that receptive fields of neurons in the horizontal kernels grow at every iteration of recurrence, and hence their range of horizontal connectivity scales positively with the amount of recurrent processing. The V1Net unit is shown in Fig.\ref{fig:v1net_arch}. 

%% file: methods_markedlong.tex
\section{Experiments}
\subsection{Perceptual grouping benchmarks -- \ml and \pf}
\begin{figure}[h]
     \centering
     \begin{subfigure}[b]{0.3\textwidth}
         \centering
        \includegraphics[width=\linewidth]{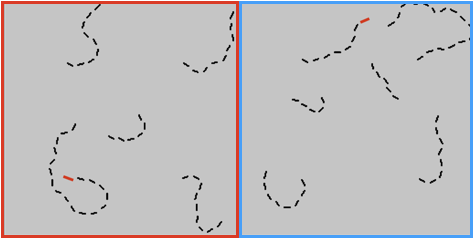}
        \caption{Samples from \ml}
         \label{fig:markedlong_samples}
     \end{subfigure}
     \hspace*{.5cm}
     \begin{subfigure}[b]{0.3\textwidth}
         \centering
        \includegraphics[width=\linewidth]{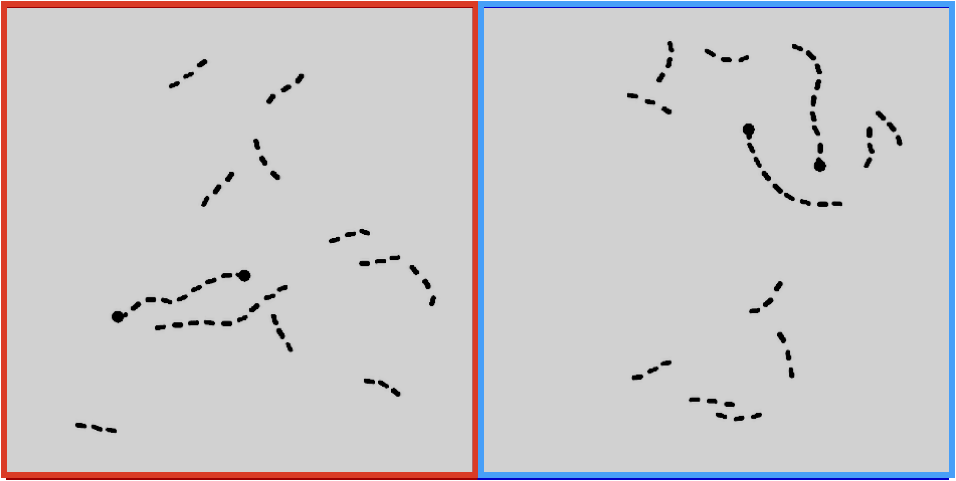}
        \caption{Samples from \pf}
         \label{fig:pf_samples}
     \end{subfigure}
     \caption{Sample images from the \ml (a) and \pf (b) datasets. Images are color-inverted for viewing convenience (keeping marked \pf segment red).}
     \label{fig:ml_sample}
    \end{figure}
\textbf{Dataset 1: \ml contour integration benchmark}
We introduce the \ml contour integration benchmark, a large scale 2-Alternative Forced Choice (2-AFC) task, i.e., a binary image classification problem that tests model ability to learn low-level serial grouping routines. 
In MarkedLong, 
each image quadrant contains one connected path $P_i$ composed of $l_i$ locally co-circular oriented segments of equal size. Three of these paths have a length $l_i=12$, while the remaining path has length $l_j=18$. One of the 54 segments of these 4 connected paths is rendered in a marker-red color, while all other segments are rendered in the foreground white color in a uniform black background. An image is classified positive if one of the segments of the long path $P_j$ is marked   red, hence the name MarkedLong; it is classified negative if one of the segments of the three short paths is marked red. 
Each path of an image is rendered with a multi-stage stochastic path-rendering algorithm discussed in the Supplementary Information (see Supplementary Sec. \ref{sec:ml_generation}) to avoid the strategy of rote-memorizing instances of the dataset. The dataset consists of 800,000 RGB images of  256 $\times$ 256 pixels. We used a 75-25 training/validation split for \ml experiments.

\textbf{Dataset 2: \pf contour integration benchmark}
We also test the models on one version of the \pf challenge introduced by \citet{linsley2018learning}. In  PathFinder, each image consists of two main connected paths $P_0$ and $P_1$ of length 9 segments as in \ml. Each image also contains two circular disks each of which is placed at one of the 4 end points of $P_0$ and $P_1$. Images that contain a disk on both ends of the same path are classified as positive, and those containing a disk on endpoints of different paths are classified as negative. This dataset consists of a total of 1,000,000 RGB images of 150 $\times$ 150 pixels. We used the medium-difficulty version of pathfinder with 9-length main paths to balance task difficulty and prototyping timescale. 
Following the experimental settings of \citet{linsley2018learning}, we used a 90-10 training/validation split for our \pf experiments. \\
Both \ml and \pf introduce short distractor paths in the background in every image to increase task difficulty via visual crowding.

%% file: experiments.tex
\subsection{Benchmarking accuracy, sample efficiency and parameter efficiency}
\begin{figure}
     \centering
     \includegraphics[width=\linewidth]{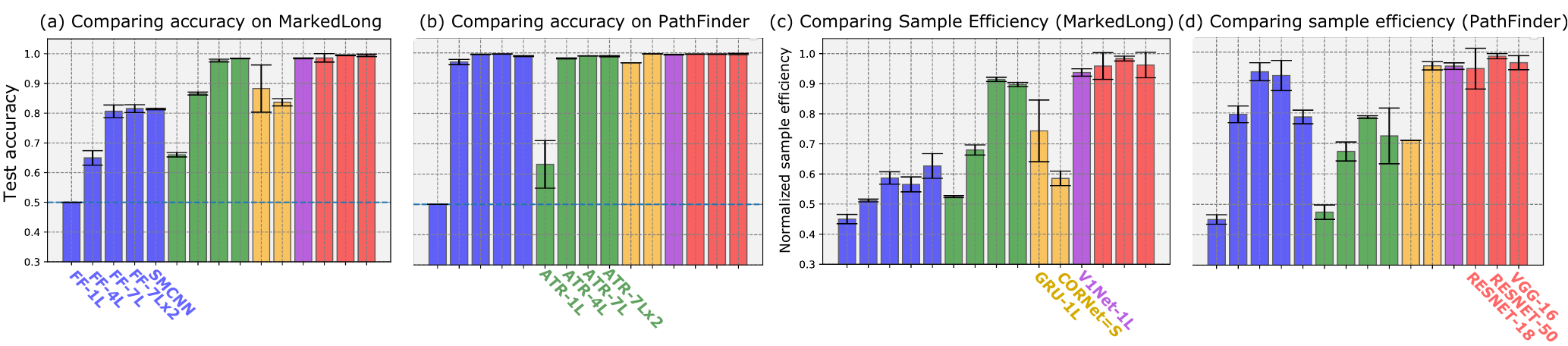} \break
     \vspace{-.3in}
     \includegraphics[width=\linewidth]{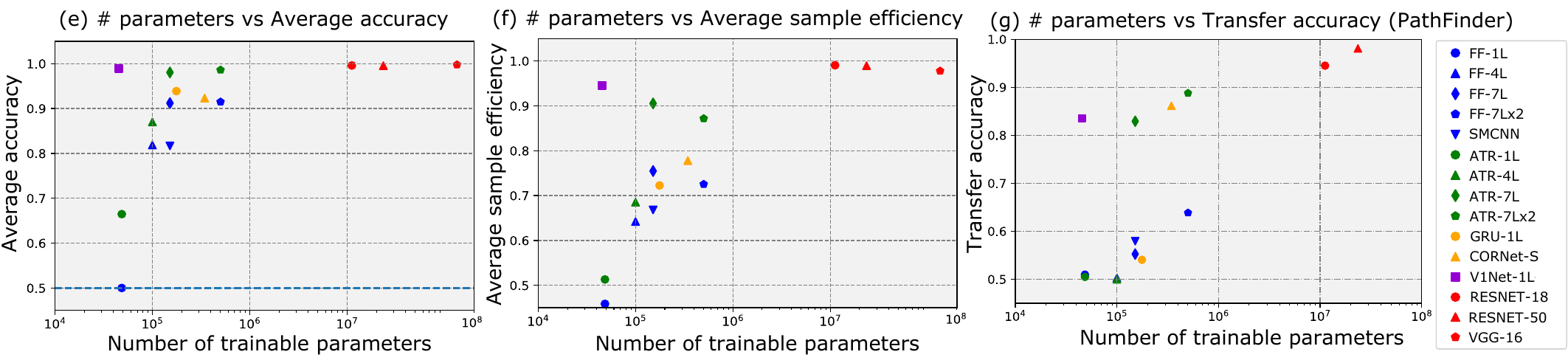}
     \caption{
      Comparing performance accuracy (a,b) and sample efficiency (c,d) on \ml (a,c) \& \pf (b,d). Parameter efficiency is computed using average accuracy (e) and sample efficiency (f) across both \ml and \pf. Transfer accuracy of models trained on \ml and finetuned on \pf shown in (g). 
     Models in families F1, F2, F3 and F4 are color-coded with blue, green, (orange, V1Net with purple) and red bars. Error bars in the figures correspond to standard deviation across 3 model runs with different random seeds. Models from F1, F2 and F3 (except V1Net) tend to learn accurate sample-efficient solutions  only on one of these two tasks as opposed to V1Nets and models in F4. V1Net's performance on-par with very-deep DCNs from F4 indicates their commensurate expressivity with DCNs which are three orders more parameter-heavy.
     }
     \label{fig:experiments}
\end{figure}

We evaluate each model using the following 3 metrics: (a) {\bf Performance accuracy}: Maximum validation accuracy measured on \ml and \pf respectively (Fig.\ref{fig:experiments}.a,b), (b) {\bf Sample efficiency}: Measure of how data-efficient learning is, computed as the area under the validation curve from beginning to the end of training on \ml and \pf respectively (Fig.\ref{fig:experiments}.c,d), and (c) {\bf Parameter efficiency}: Measured by counting the number of trainable parameters in a model.

\label{sec:benchmark}
We study 15 
different models (for exact architecture, see Supplementary Sec. \ref{sec:architectures_si}). Broadly, we classify them as belonging to the following four families of DCNs: {\bf (F1)} Standard Convolutional DCNs, {\bf (F2)} Atrous-convolutional DCNs, {\bf (F3)} Recurrent-Convolutional DCNs, and {\bf (F4)} very deep over-parameterized DCNs. 

\textbf{Implementation details} All standard convolutional DCNs (F1), Atrous-convolutional DCNs (F2) and Recurrent convolutional DCNs (F3) were implemented using a common three-block structure. These models consist of a standard \textit{input block} designed as a 7x7 convolution layer with 32 filters. A subsequent block that we denote as the \textit{intermediate block} is unique to each model and responsible for learning task-relevant grouping transformations of the input activity. Our \textit{readout block} consisted of a Global-Average-Pooling operation followed by a 512-D fully connected layer and a 2-D output layer. 
Parameters were initialized with the Variance Scaling method \cite{he2015delving} and optimized using the Adam optimizer \cite{Kingma2015Adam:Optimization} with initial learning rate of 1e-3 on Pathfinder and 5e-4 on MarkedLong respectively, $\beta_1=0.9$, $\beta_2=0.999$ and $\epsilon=1e-8$. All models were implemented using the TensorFlow framework \cite{abadi2016tensorflow} and trained on 8-core Google Cloud TPU-v3 nodes with a batch size of 128.

\textbf{F1: Standard Convolutional DCNs}:
We compare 5 standard DCN models with varying amounts of hierarchical processing. Our comparison consists of (1) a 3-layer DCN (FF-1L), (2) a 5-layer DCN (FF-4L) and (3) a 7-layer DCN (FF-7L). To assess the impact of the number of output convolutional kernels at a given depth of processing, we evaluated (4) FF-7Lx2, a 7-layer DCN in which each intermediate block convolutional layer learns twice the number of filters as in FF-7L. Intermediate block convolution layers of (1), (2) and (3) contained a filter bank of 32 5x5 kernels, while (4) contained 64 5x5 kernels.  We implemented (5) a feedforward surround-modulation CNN 
\cite{Hasani2019SurroundNetworks}
to assess the role of feedforward vs recurrent lateral connections in grouping. FF-SMCNN's architecture is identical to that of FF-7L, 
but its input convolution layer consists of fixed DoG kernels (in addition to trainable kernels) that implement convolutional surround modulation. None of these tested feedforward DCNs matched the accuracy  of our proposed V1Net model on MarkedLong and had a strikingly large sample efficiency difference compared to V1Net. This observation suggests an inability of feedforward DCNs in F1 to efficiently group long paths in \ml.
On PathFinder, only the deep 7-layer FF-7L and FF-7L-x2 matched V1Net's performance. DCNs in F1 which, at their best come close to matching V1Net's performance, are significantly more parameter-expensive.

\textbf{F2: Atrous-convolutional DCNs}: 
An exponential growth in the receptive field size of dilated convolution operations, or atrous convolutions, have been shown to
be useful for  learning  long-range contextual features in DCNs 
and should hence be beneficial for our contour integration tasks.
We compare 4 different atrous convolutional DCNs with varying amount of processing depth. We implement (6) a 3-layer atrous-DCN (ATR-1L), (7) a 5-layer atrous-DCN (ATR-4L), (8) a 7-layer atrous-DCN (ATR-7L) and (9) a high parameter variant of ATR-7L with twice the convolutional filters at each intermediate block convolutional layer that we denote as ATR-7Lx2. We implemented these models by replacing the standard 2D-convolution layers in models (1,2,3,4) with atrous convolution layers \cite{Chollet2017Xception:Convolutions} while maintaining the kernel size as 5x5 within the intermediate block. Among these models, we note that the shallow Atrous-DCNs are significantly worse than V1Net, i.e, our experiments found Atrous Convolutions to be useful only when they were accompanied by a deep hierarchical architecture (as in ATR-7L, ATR-7Lx2). 
However, 
they struggle to learn the PathFinder challenge wherein they fail to match the sample efficiency of V1Net by a significant extent. 

\textbf{F3: Recurrent convolutional DCNs}: 
We test the utility of recurrent computations by evaluating 3 recurrent-convolutional DCNs with varying inspiration and architecture. 
Each of these networks is designed as a 3-layer DCN with the intermediate block comprised of one of our chosen recurrent-convolutional layers with 5 iterations of recurrence. In this class, we evaluate (10) GRU-1L \cite{cho2014learning} wherein the intermediate block consists of a Convolutional-GRU cell, (11) CORNet-S wherein the intermediate block consists of the bio-inspired CORNet-S module \cite{kubilius2018cornet},
(12) V1Net-1L wherein the middle layer consists of a V1Net module.
Results from this family of models suggest that not all recurrent architectures perform equally. CORNet-S, inspired by the concepts of weight sharing and residual skip connections matches V1Net's accuracy and sample efficiency on \pf. However on \ml, CORNet-S has a significantly poor sample efficiency compared to V1Net. GRU-1L matches the accuracy of V1Net on both \ml and \pf, however, it has significantly worse sample efficiency on both problems. 
These models implement recurrent computations and consistently achieve high accuracies, suggesting they are learning grouping routines. However, they lack the diverse types of recurrent horizontal connections, gain modulation mechanism, and various cell-types that V1Net additionally possesses. We hypothesize this to be the reason for their sub-optimal learning speed compared to V1Net. 
While both GRU-1L and CORNet-S contain more parameters than V1Net, they are relatively more parameter-efficient compared to FF-7Lx2 models and those in F4 discussed next. 

\textbf{F4: Very deep overparameterized DCNs}:
We evaluate three previously published state-of-the-art DCNs which contain about $10^3 \times$ more trainable parameters than a V1Net-1L model. In this class, we study the performance of (13) a variant of the VGG-16 model \cite{simonyan2014very} with a batch normalization layer before every nonlinear rectification layer, (14) the latest  ResNet-18 model with pre-activation \cite{he2016deep}, and (15) the latest ResNet-50 model with pre-activation. We report experiments wherein we trained these models from scratch as we noticed that ImageNet-pretraining was inferior
due to a significant domain shift. All models we tested in this family of over-parametrized DCNs showed consistently high performance on both tasks, similar to V1Net's performance.

We highlight that V1Net-1L, while being the most parameter efficient model we compared, consistently beats the accuracy and sample efficiency of all 11 models in the first three model families F1, F2 and F3 on \pf and \ml. Negligible differences in the performance of the V1Net with those in F4 suggest that the extremely compact V1Net representations are equally as expressive as that of the evaluated very-deep state-of-the-art DCNs with just a fraction of trainable weights as is observed in Fig. \ref{fig:experiments}.e,f. comparing model parameter efficiency.
\subsection{V1Net learns the most compact generalizable grouping routines}
Although \ml  and \pf are different problems, we recognize a latent similarity between them in that generalizable solutions to either task involve similar functional aspects relating to contour integration. 
To investigate this, we explore the transfer learning accuracy of each model from \ml to  \pf.
We trained each model 
on \ml and finetuned the output fully-connected layers (for learning a new readout strategy) and all batch-normalization parameters (including ones in the input and intermediate blocks - to adapt to \pf's minibatch statistics) on \pf. We froze all convolution layers in the input and intermediate blocks (that we consider to implement contour grouping routines learned on \ml).
Our observations from this experiment show strong evidence supporting our initial hypothesis that carefully designed recurrent architectures learn compact generalizable solutions to perceptual grouping. Among the 15 models compared on this transfer learning experiment, only half the models showed promising generalization performance (Fig.\ref{fig:experiments}.g). None of the standard DCN models from F1 generalize to \pf; Fig.\ref{fig:experiments}.e shows that the majority of these models (except shallow 1-layer DCNs which show poor source performance) fail to generalize despite learning successful solutions to solve \ml. Only the deep Atrous-DCNs from F2 show good generalization, further highlighting their dependency on parameter-heavy solutions and relatively more hierarchical processing compared to recurrent architectures. Out of the 3 tested recurrent architectures in F3, our proposed V1Net and CORNet-S generalize successfully. All overparameterized DCNs from F4 generalize well to \pf. We are ideally looking for compact models with high parameter efficiency and transfer accuracy which lie on the upper left-hand corner of Fig. \ref{fig:experiments}.g., it has to be noted that V1Nets are the closest models in this region and are finetuning many fewer variables (10 variables) relative to CORNet-S (70) and DCNs in F4(200) that generalize well. 
\subsection{Interpreting V1Net's \ml grouping strategy}
\label{sec:grouping_vis}
\begin{figure}[]
\begin{subfigure}[b]{\linewidth}
\centering
    \includegraphics[width=\linewidth]{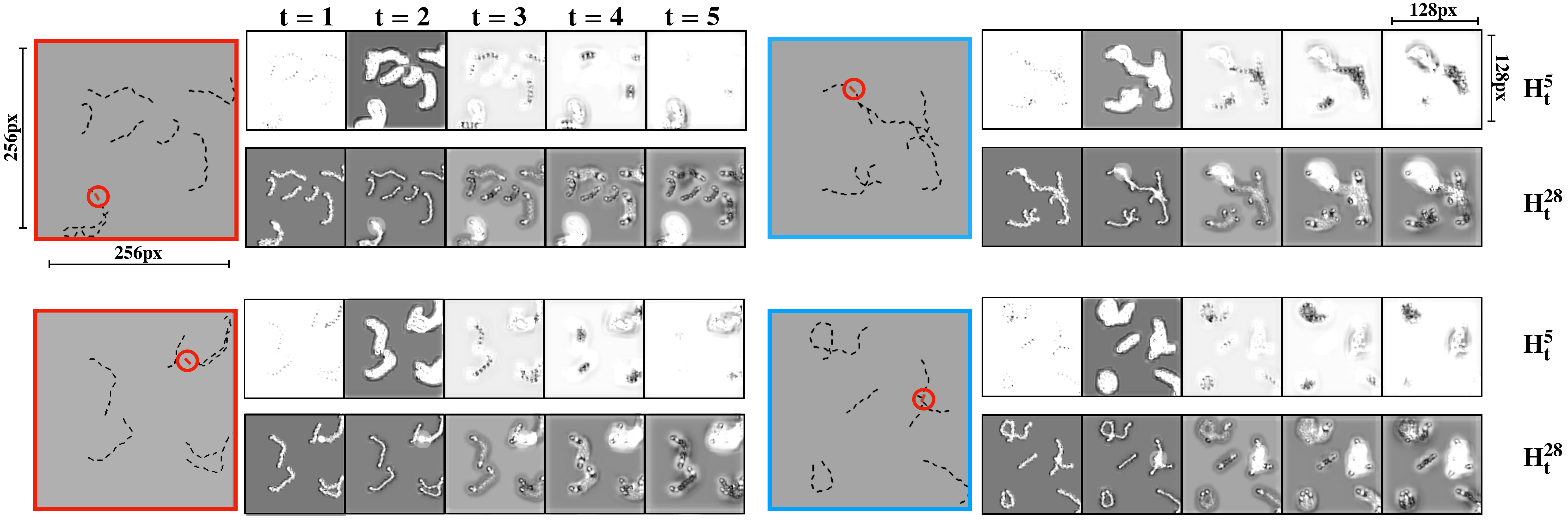}
    \caption{Evolution of V1Net's activations through recurrent iterations}
    \label{fig:v1net_activations}
\end{subfigure}
\break 
\begin{subfigure}[b]{\linewidth}
    \centering
    \includegraphics[width=\linewidth]{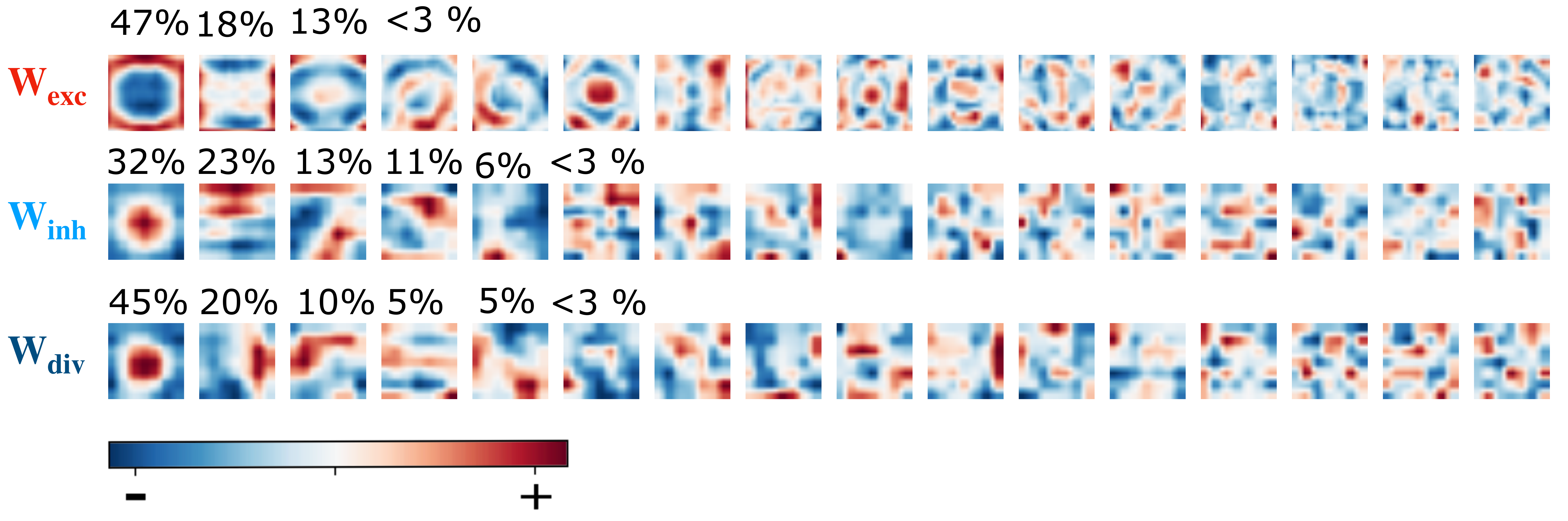}
    \caption{Representative horizontal connections learned in V1Net}
    \label{fig:pca_horizontal}
\end{subfigure}
    \caption{(a) Visualizing temporal dynamics of V1Net activations for \ml. Marker locations highlighted with red circles for viewing convenience. (b) Visualizing the top PCs computed from a bank of excitatory (top), inhibitory (middle) and divisive inhibition (bottom) horizontal connection kernels learned by a V1Net trained on \ml sorted in decreasing order of their ratio of explained variance (mentioned in figure only for top PCs). See Sec. \ref{sec:grouping_vis}}
    \vspace*{-.15in}
    \label{fig:interpretability}
\end{figure}

In order to qualitatively understand V1Net's solution to \ml, we visualized the temporal evolution of activation maps in $\htt$, the internal state of a trained V1Net model that reflects grouping on \ml. In Fig. \ref{fig:v1net_activations}, we show a subset of these visualizations on a couple of positive (left column) and negative (right column) class images from \ml: the $5^{th} (\mbf{H^5_t})$ and $28^{th} (\mbf{H^{28}_t})$ activation maps from a V1Net trained for 5 recurrent iterations. Each column of activation maps corresponds to the model's response at a particular timestep of processing. 
(1) Analyzing activations from $\mbf{H^{28}_t}$ suggests that the activity along every path in the input gets progressively enhanced. Particularly, a blob of activity marking the 'red marker' is incrementally enhanced through recurrent processing to create an attention map focused on the marked path. 
(2) Activations from $\mbf{H^5_t}$ are equally informative. 
During early recurrent steps, all paths are activated sparsely in $\mbf{H^5_t}$. With more recurrent processing, only the marked path (bright blob in Fig. \ref{fig:v1net_activations}) and longest path (dark blob) get enhanced activity consistently as they both compete and suppress the activity of all remaining paths. 
The top-right panel of Fig. \ref{fig:v1net_activations}. is a particularly difficult example for even humans to quickly parse the grouping of the marked and long paths. It is interesting how in early recurrent steps, similar to the human visual system, V1Net groups together elements of two paths that intersect but later through further recurrent processing infers the correct part-whole ownership via long-range contextual feature integration. It then separates the long and marked path components into their respective light/dark groups. These visualizations provide an interpretation of how V1Net combines different algorithmic grouping blocks (attending to the marker, growing paths to find the longest path, etc). 

We examined V1Net's learned horizontal connection patterns on \ml by applying PCA to a bank of 32 excitatory ($\wexc$), inhibitory ($\winh$) and divisive inhibition ($\wdiv$) horizontal connection kernels each (Fig. \ref{fig:pca_horizontal}.). The resulting top 4 principal components explained about 80\% of the total variance in each of these filter banks. These highly represented PC's -- that resemble suppressive center-surround and facilitatory 'association-field' interactions -- bear similar structure to biological horizontal connections in primate area V1 \cite{Gilbert1989ColumnarCortex,Field1993ContourField}. Further PCs are likely modeling intricate higher-order interactions.

%% file: discussion.tex
\section{Discussion}
We hypothesized that recurrence-infused DCNs are capable of learning superior solutions to perceptual grouping problems involving contour integration governed by low-level Gestalt cues. We test this hypothesis through a two-pronged approach and find evidence in support of our hypothesis: A 3-layer V1Net-DCN learns the most compact, generalizable solutions that reach near-perfect accuracy solutions to both \ml and \pf challenges with exceptional sample efficiency that is comparable only to very-deep and parameter-heavy state of the art DCNs with three orders of magnitude more trainable parameters among a broad range of 15 feedforward and recurrent models. Visualizing the internal state of V1Net's recurrent grouping routine provides an algorithmic interpretation linking how V1Net's low-level recurrent horizontal interactions achieve the high-level computational goals of \ml and \pf.

Our experimental results accord with prior findings from neuroscience and vision science on how the ventral visual stream performs perceptual grouping effortlessly using recurrent interactions despite their relatively shallow anatomy (compared to recent DCNs with multiple orders greater hierarchy) \cite{Lamme1998FeedforwardCortex,Roelfsema2006CORTICALGROUPING,Kreiman2020BeyondCortex}. Here, we show that standard DCNs struggle to match the superior performance, sample efficiency and generalizability of the proposed recurrent V1Net-1L model on 2 contour integration tasks. 
On this note we believe that the utility of bio-inspired recurrent models such as V1Net should be explored in 
overcoming some of the 
recent deficiencies (where texture is biased over shape) observed in DCNs \cite{geirhos2018imagenettrained,Brendel2019ApproximatingImageNet,Baker2018DeepShape,long2018role,Laskar2018CorrespondenceTextures} by emphasizing extended contour processing through 
recurrent horizontal interactions  \cite{Loffler2008PerceptionMechanisms,Elder2018ShapeRepresentation,Elder2018TheShapes}.


On the other hand, we are intrigued by the superior ability of parameter-heavy feedforward state-of-the-art DCNs in performing perceptual grouping with identical performance to recurrent DCNs. Further analyses of strategies learned by these models will inform future studies of cortical processing to search for currently unexplored mechanisms achievable by nuanced feedforward computations.

In summary, our work highlights the efficacy of recurrent computations in learning highly compact, generalizable solutions to perceptual grouping problems such as the proposed \ml and \pf challenges. Our bio-inspired V1Net recurrent unit learns the most compact generalizable solutions to the above grouping problems, and it evolves interpretable internal states reflecting the model's learned low-level grouping strategies. These characteristics of V1Net make it suitable  for  sample- and parameter-efficient on-device computer vision as well as computational modeling of primate cortical processing.

%% file: acknowledgments.tex
\section*{Acknowledgments}
We thank Drew Linsley and Thomas Serre for sharing with us the PathFinder challenge dataset.
This work was supported by the Kavli Symposium Inspired Proposals award from the Kavli Institute of Brain and Mind, the Department of Cognitive Science at UC San Diego, Cloud TPU resources made available by Google through the TensorFlow Research Cloud program, Social Sciences Computing Facility at UC San Diego,  Nautilus compute cluster supported by Pacific Research Platform, and the Titan V GPU donated by NVIDIA Corporation.

%% file: neurips_2020_si.tex
\renewcommand{\thesection}{S\arabic{section}}
\renewcommand{\thetable}{S\arabic{table}}
\renewcommand{\thefigure}{S\arabic{figure}}
\newpage
\setcounter{section}{0}
\section*{Supplementary information}
\section{Architecture details of feedforward and recurrent comparison models}
\label{sec:architectures_si}
All models from F1, F2, and F3 were implemented using the three-block structure mentioned in the main paper (Input block, intermediate block, readout block). While the input and readout blocks were fixed to be standard across these three families, each model had a distinct architecture as its intermediate block. For explanation convenience, we merge the two dense layers in our readout block into a single readout layer. Architectural details for all these models are available in Tables. \ref{tab:ff_architectures},\ref{tab:atrous_architectures},\ref{tab:rcnn_architectures}. 
\subsection*{Family 1: Standard Convolutional DCNs}
The intermediate layers of all F1 models contain standard 2D convolutions implemented in the following TensorFlow call: tensorflow.nn.conv2D()
While the intermediate block's standard convolution filter banks (models 1-3) contain 32 output filters, wide convolution filter bank (model 4) contains 64 output filters. Our FF-SMCNN model implementation of \citet{} Conv2D\_SM input layer. Conv2D\_SM is an implementation of \citet{Hasani2019SurroundNetworks} wherein one half of the convolutional filter bank consists of non-trainable Difference-of-Gaussian filters that implement a convolutional surround modulation of the bottom-up response. The input convolution layer is followed by a max-pooling operation, all convolutions in F1 use 'SAME' padding and weights are initialized using Variance Scaling method. Details of each model's architecture is available in Table.\ref{tab:ff_architectures}. 

\subsection*{Family 2: Atrous Convolutional DCNs}
All intermediate layers in F2 use atrous convolutions / dilated convolutions with a dilation factor of 2. This layer is implemented in the following tensorflow call: tensorflow.nn.atrous\_conv2d. The input convolution layer is followed by a max-pooling operation, all convolutions in F1 use 'SAME' padding and weights are initialized using Variance Scaling method. Details of each model's architecture is available in Table.\ref{tab:atrous_architectures}.

\subsection*{Family 3: Recurrent Convolutional DCNs}
We tested 3 recurrent convolutional DCNs (10) GRU-1L, a three layer DCN with a 1 layer intermediate block consisting of a Convolutional-GRU cell (we used a TensorFlow implementation based on the ConvGRU cell available in 
\url{https://github.com/carlthome/tensorflow-convlstm-cell}
, (11) CORNet-S, a 3 layer DCN with a 1 layer intermediate block consisting of the CORNet-S block from \citet{kubilius2018cornet} (we used a TensorFlow implementation based on 
\url{https://github.com/dicarlolab/CORnet/blob/master/cornet/cornet_s.py}
)
, and (12) V1Net-1L, a 3 layer DCN with a 1 layer intermediate block consisting of our proposed V1Net architecture. More details on each model's architecture are available in Table.\ref{tab:rcnn_architectures}.

\subsection*{Family 4: Very deep overparameterized DCNs}
For this model family, we tested (13) a implementation of VGG-16 with batch normalization before each ReLU layer, (14) the official TensorFlow implementation of ResNet-v2 (18 layers deep) and (16) the official TensorFlow implementation of ResNet-v2 (50 layers deep). These ResNet-v2 implementations are available at 
\url{https://github.com/tensorflow/tpu/tree/master/models/official/resnet}

\section*{V1Net architecture details and discussion}
V1Net models in our experiments were trained with 5 recurrent iterations. The kernel width of excitatory connections $\wexc$ is [15,15], which is thrice the size of our input kernel $\wxh$ in order to support far long-range excitatory connections. Inhibitory and divisive gain control connections are implemented by [7,7] separable convolution kernels $\winh$ and $\wdiv$ respectively. $\winh$ and $\wdiv$ model short-range linear and nonlinear contextual competitive interactions of bottom up representations. 

Each of these three filter banks above learn $k$ canonical spatial templates (such as center-on surround-off, horizontally elongated association fields, etc.) for supporting excitatory and inhibitory interactions between $k^2$ pairs of bottom-up features. These spatial templates are used in combination with pair-wise weights learned by pointwise convolutions to facilitate spatially selective pairwise interactions between $k^2$ input feature pairs.
We see depthwise separable convolutional horizontal connections as a special case of vanilla 2D convolutional horizontal connections. While 2D convolutional horizontal connections learn distinct spatial profiles and their corresponding feature combinations jointly, separable convolutional horizontal connections factorize this computation into canonical spatial profile learning (depthwise convolution) and feature combination weight learning (pointwise convolution).

\section{\ml generation:}
\label{sec:ml_generation}
The central component to generating \ml images is our stochastic path-rendering algorithm described below. As mentioned in the main paper, all \ml images contain exactly one long path (18 bars long) and three short paths (12 bars long). In addition to these 4 paths, few more shorter distractor paths (6 bars long) are placed at randomly sampled locations within the image. Each one of the 4 primary paths (18- and 12-bars long) originates from a seed location sampled uniformly within one of the 4 quadrants in an image (each path seeds from a unique quadrant). Deliberately seeding paths in unique quadrants reduces the amount of intersection (or) overlap of different paths by pushing them away from each other. Originating from these seed locations, fully grown connected paths of desired lengths are developed using the following iterative algorithm.
\begin{figure}[h]
    \centering
    \includegraphics[width=\linewidth]{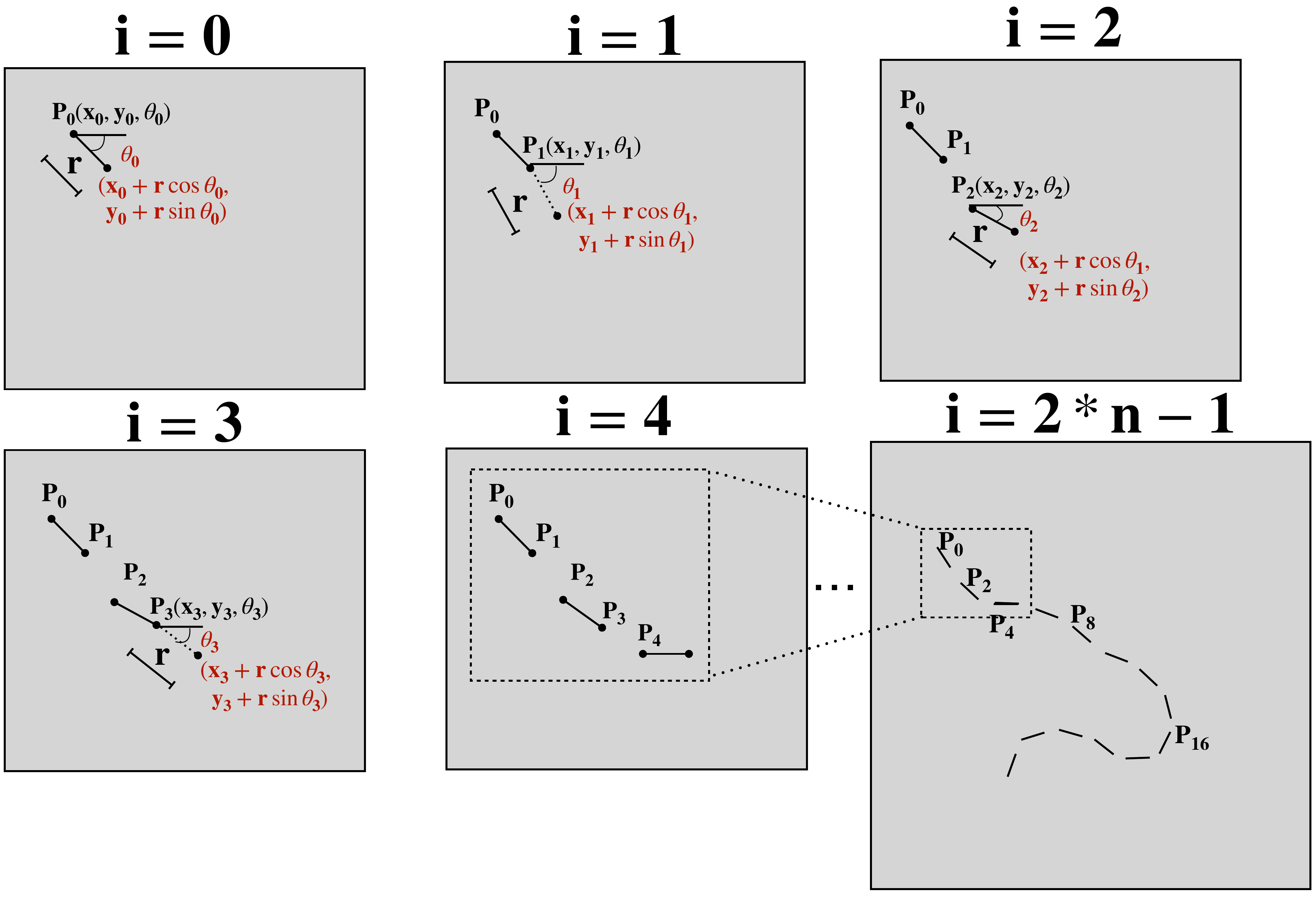}
    \caption{Step by step illustration of \ml's path generation algorithm. The shown bar sizes / distance between bars in this figure is not representative of images in our dataset, we use the most suitable settings for reader's convenience in this figure.}
    \label{fig:ml_generation}
\end{figure}

\textbf{Path rendering description:}
Our algorithm samples 2*n points that lie on a smooth connected curve that is locally co-circular; every alternate pair of points is then connected to form an n-length path.
A seed pixel location $\mbf{P_0(x_0,y_0)}$ is sampled uniformly within a desired quadrant's boundaries as the starting point of the first bar. The orientation $\mbf{\theta_0}$ of this bar is sampled from the uniform distribution $\mbf{\theta \sim U(0, \pi/4)}$. The end point of this bar, which acts as the starting point for the next iteration, is computed by projecting a vector of length $\mbf{r}$ at an angle $\theta_0$ to the horizontal axis, $\mbf{P_1(x_1,y_1) = (x_0+r\cos{\theta_0},y_0+r\sin{\theta_0})}$. 

Each successive point $\mbf{P_{i+1}}$ is updated to $\mbf{[x_i+r\cos{\theta_{i+1}}, y_i+r\sin{\theta_{i+1}}]}$, $\mbf{\theta_{i+1}}$ is sampled from the uniform distribution  $\theta \sim U(\theta_i-\pi/4, \theta_i+\pi/4)$. In the event of $\mbf{P_{i+1}}$ falling outside the image boundary, the algorithm backtracks two steps to step $i-1$, and continues rendering along a different path. This iterative process stops upon reaching the exit condition, $\mbf{i=2*n-1}$. Every pair of adjacent points $\mbf{P_{2*i},P_{2*i+1}, i \in [0,n-1]}$ are connected using line segments at this stage to render a connected n-length path. A pictorial walkthrough of this algorithm is shown in Fig. \ref{fig:ml_generation}.

\textbf{Generating positive class images:}
We sample one quadrant $i_q^l \in [0,3]$ within which the longest path is seeded. Path rendering algorithm selects a seed within this quadrant and grows a path of length $n=18$. A marker index $i_m$ is sampled uniformly from $i \sim U[0,17]$ and rendered with a line segment red in color. Following longest marked path's generation, one short path is seeded and generated in each remaining quadrant with length $n=12$. $n_d \sim U[1,4]$ number of distractors are then rendered within the entire image.

\textbf{Generating negative class images:}
In order to generate negative class images, none of the bars in the longest path seeded in $i_q^l$ are marked with a red marker bar. Instead, one of the three 12-length bars is chosen at random to be marked; a marker index $i_m$ is sampled uniformly from $i \sim U[0,11]$ and this bar on the chosen short path is rendered with a line segment red in color.

In total, \ml consists of 400,000 positive class images and 400,000 negative class images; 100,000 images rendered in each class are flipped horizontally and vertically through online data augmentation, resulting in 400,000 images per class. 

See next page for architecture detail tables.
 
\newpage
\section*{Tables with architectural details of comparison models}
\begin{table}[h]
\caption{Feedforward architecture details}

\begin{tabular}{|l|l|l|l|l|l|l|}
\hline
\multicolumn{7}{|l|}{\textbf{FF-1L}}                                                                                        \\ \hline
Layer              & Layer type & Kernel size   & N\_out & Pooling & Input            & Output           \\ \hline
1                  & Conv2D     & {[}7,7{]}     & 32               & Max     & {[}H,W,3{]}  & {[}H/2,W/2,32{]} \\ \hline
2                & Conv2D     & {[}5,5{]}     & 32               & None    & {[}H/2,W/2,32{]} & {[}H/2,W/2,32{]} \\ \hline
\multirow{4}{*}{3} & GAP        & {[}H/2,W/2{]} & -                & None    & {[}H/2,W/2,32{]} & {[}1,1,32{]}     \\ \cline{2-7} 
                   & Dense      & -             & 512              & None    & {[}32{]}         & {[}512{]}        \\ \cline{2-7} 
                   & Dense      & -             & 2                & None    & {[}512{]}        & {[}2{]}          \\ \cline{2-7} 
                   & Softmax    & -             & 2                & None    & {[}2{]}          & {[}2{]}          \\ \hline
\end{tabular}
\begin{tabular}{|l|l|l|l|l|l|l|}
\hline
\multicolumn{7}{|l|}{\textbf{FF-4L}}                                                                                        \\ \hline
Layer              & Layer type & Kernel size   & N\_out & Pooling & Input            & Output           \\ \hline
1                  & Conv2D     & {[}7,7{]}     & 32               & Max     & {[}H,W,3{]}  & {[}H/2,W/2,32{]} \\ \hline
2-4                & Conv2D     & {[}5,5{]}     & 32               & None    & {[}H/2,W/2,32{]} & {[}H/2,W/2,32{]} \\ \hline
\multirow{4}{*}{5} & GAP        & {[}H/2,W/2{]} & -                & None    & {[}H/2,W/2,32{]} & {[}1,1,32{]}     \\ \cline{2-7} 
                   & Dense      & -             & 512              & None    & {[}32{]}         & {[}512{]}        \\ \cline{2-7} 
                   & Dense      & -             & 2                & None    & {[}512{]}        & {[}2{]}          \\ \cline{2-7} 
                   & Softmax    & -             & 2                & None    & {[}2{]}          & {[}2{]}          \\ \hline
\end{tabular}
\begin{tabular}{|l|l|l|l|l|l|l|}
\hline
\multicolumn{7}{|l|}{\textbf{FF-7L}}                                                                                        \\ \hline
Layer              & Layer type & Kernel size   & N\_out & Pooling & Input            & Output           \\ \hline
1                  & Conv2D     & {[}7,7{]}     & 32               & Max     & {[}H,W,3{]}  & {[}H/2,W/2,32{]} \\ \hline
2-6                & Conv2D     & {[}5,5{]}     & 32               & None    & {[}H/2,W/2,32{]} & {[}H/2,W/2,32{]} \\ \hline
\multirow{4}{*}{7} & GAP        & {[}H/2,W/2{]} & -                & None    & {[}H/2,W/2,32{]} & {[}1,1,32{]}     \\ \cline{2-7} 
                   & Dense      & -             & 512              & None    & {[}32{]}         & {[}512{]}        \\ \cline{2-7} 
                   & Dense      & -             & 2                & None    & {[}512{]}        & {[}2{]}          \\ \cline{2-7} 
                   & Softmax    & -             & 2                & None    & {[}2{]}          & {[}2{]}          \\ \hline
\end{tabular}

\begin{tabular}{|l|l|l|l|l|l|l|}
\hline
\multicolumn{7}{|l|}{\textbf{FF-7L-x2}}                                                                                        \\ \hline
Layer              & Layer type & Kernel size   & N\_out & Pooling & Input            & Output           \\ \hline
1                  & Conv2D     & {[}7,7{]}     & 32               & Max     & {[}H,W,3{]}  & {[}H/2,W/2,32{]} \\ \hline
2-6                & Conv2D     & {[}5,5{]}     & 64               & None    & {[}H/2,W/2,32{]} & {[}H/2,W/2,64{]} \\ \hline
\multirow{4}{*}{7} & GAP        & {[}H/2,W/2{]} & -                & None    & {[}H/2,W/2,64{]} & {[}1,1,64{]}     \\ \cline{2-7} 
                   & Dense      & -             & 512              & None    & {[}64{]}         & {[}512{]}        \\ \cline{2-7} 
                   & Dense      & -             & 2                & None    & {[}512{]}        & {[}2{]}          \\ \cline{2-7} 
                   & Softmax    & -             & 2                & None    & {[}2{]}          & {[}2{]}          \\ \hline
\end{tabular}

\begin{tabular}{|l|l|l|l|l|l|l|}
\hline
\multicolumn{7}{|l|}{\textbf{FF-7L-SMCNN}}                                                                                        \\ \hline
Layer              & Layer type & Kernel size   & N\_out & Pooling & Input            & Output           \\ \hline
1                  & Conv2D\_SM     & {[}7,7{]}     & 32               & Max     & {[}H,W,3{]}  & {[}H/2,W/2,32{]} \\ \hline
2-6                & Conv2D     & {[}5,5{]}     & 32               & None    & {[}H/2,W/2,32{]} & {[}H/2,W/2,32{]} \\ \hline
\multirow{4}{*}{7} & GAP        & {[}H/2,W/2{]} & -                & None    & {[}H/2,W/2,32{]} & {[}1,1,32{]}     \\ \cline{2-7} 
                   & Dense      & -             & 512              & None    & {[}32{]}         & {[}512{]}        \\ \cline{2-7} 
                   & Dense      & -             & 2                & None    & {[}512{]}        & {[}2{]}          \\ \cline{2-7} 
                   & Softmax    & -             & 2                & None    & {[}2{]}          & {[}2{]}          \\ \hline
\end{tabular}

\label{tab:ff_architectures}
\end{table} 

\begin{table}[h]
\caption{Atrous convolutional DCN architecture details}

\begin{tabular}{|l|l|l|l|l|l|l|}
\hline
\multicolumn{7}{|l|}{\textbf{ATR-1L}}                                                                                        \\ \hline
Layer              & Layer type & Kernel size   & N\_out & Pooling & Input            & Output           \\ \hline
1                  & Conv2D     & {[}7,7{]}     & 32               & Max     & {[}H,W,3{]}  & {[}H/2,W/2,32{]} \\ \hline
2                & Atrous\_Conv2D     & {[}5,5{]}     & 32               & None    & {[}H/2,W/2,32{]} & {[}H/2,W/2,32{]} \\ \hline
\multirow{4}{*}{3} & GAP        & {[}H/2,W/2{]} & -                & None    & {[}H/2,W/2,32{]} & {[}1,1,32{]}     \\ \cline{2-7} 
                   & Dense      & -             & 512              & None    & {[}32{]}         & {[}512{]}        \\ \cline{2-7} 
                   & Dense      & -             & 2                & None    & {[}512{]}        & {[}2{]}          \\ \cline{2-7} 
                   & Softmax    & -             & 2                & None    & {[}2{]}          & {[}2{]}          \\ \hline
\end{tabular}

\begin{tabular}{|l|l|l|l|l|l|l|}
\hline
\multicolumn{7}{|l|}{\textbf{ATR-4L}}                                                                                        \\ \hline
Layer              & Layer type & Kernel size   & N\_out & Pooling & Input            & Output           \\ \hline
1                  & Conv2D     & {[}7,7{]}     & 32               & Max     & {[}H,W,3{]}  & {[}H/2,W/2,32{]} \\ \hline
2-4                & Atrous\_Conv2D     & {[}5,5{]}     & 32               & None    & {[}H/2,W/2,32{]} & {[}H/2,W/2,32{]} \\ \hline
\multirow{4}{*}{5} & GAP        & {[}H/2,W/2{]} & -                & None    & {[}H/2,W/2,32{]} & {[}1,1,32{]}     \\ \cline{2-7} 
                   & Dense      & -             & 512              & None    & {[}32{]}         & {[}512{]}        \\ \cline{2-7} 
                   & Dense      & -             & 2                & None    & {[}512{]}        & {[}2{]}          \\ \cline{2-7} 
                   & Softmax    & -             & 2                & None    & {[}2{]}          & {[}2{]}          \\ \hline
\end{tabular}
\begin{tabular}{|l|l|l|l|l|l|l|}
\hline
\multicolumn{7}{|l|}{\textbf{ATR-7L}}                                                                                        \\ \hline
Layer              & Layer type & Kernel size   & N\_out & Pooling & Input            & Output           \\ \hline
1                  & Conv2D     & {[}7,7{]}     & 32               & Max     & {[}H,W,3{]}  & {[}H/2,W/2,32{]} \\ \hline
2-6                & Atrous\_Conv2D     & {[}5,5{]}     & 32               & None    & {[}H/2,W/2,32{]} & {[}H/2,W/2,32{]} \\ \hline
\multirow{4}{*}{7} & GAP        & {[}H/2,W/2{]} & -                & None    & {[}H/2,W/2,32{]} & {[}1,1,32{]}     \\ \cline{2-7} 
                   & Dense      & -             & 512              & None    & {[}32{]}         & {[}512{]}        \\ \cline{2-7} 
                   & Dense      & -             & 2                & None    & {[}512{]}        & {[}2{]}          \\ \cline{2-7} 
                   & Softmax    & -             & 2                & None    & {[}2{]}          & {[}2{]}          \\ \hline
\end{tabular}

\begin{tabular}{|l|l|l|l|l|l|l|}
\hline
\multicolumn{7}{|l|}{\textbf{ATR-7L-x2}}                                                                                        \\ \hline
Layer              & Layer type & Kernel size   & N\_out & Pooling & Input            & Output           \\ \hline
1                  & Conv2D     & {[}7,7{]}     & 32               & Max     & {[}H,W,3{]}  & {[}H/2,W/2,32{]} \\ \hline
2-6                & Atrous\_Conv2D     & {[}5,5{]}     & 64               & None    & {[}H/2,W/2,32{]} & {[}H/2,W/2,64{]} \\ \hline
\multirow{4}{*}{7} & GAP        & {[}H/2,W/2{]} & -                & None    & {[}H/2,W/2,64{]} & {[}1,1,64{]}     \\ \cline{2-7} 
                   & Dense      & -             & 512              & None    & {[}64{]}         & {[}512{]}        \\ \cline{2-7} 
                   & Dense      & -             & 2                & None    & {[}512{]}        & {[}2{]}          \\ \cline{2-7} 
                   & Softmax    & -             & 2                & None    & {[}2{]}          & {[}2{]}          \\ \hline
\end{tabular}
\label{tab:atrous_architectures}
\end{table} 

\begin{table}[h]
\caption{Recurrent convolutional DCN architecture details}
\begin{tabular}{|l|l|l|l|l|l|l|}
\hline
\multicolumn{7}{|l|}{\textbf{GRU-1L}}                                                                                        \\ \hline
Layer              & Layer type & Kernel size   & N\_out & Pooling & Input            & Output           \\ \hline
1                  & Conv2D     & {[}7,7{]}     & 32               & Max     & {[}H,W,3{]}  & {[}H/2,W/2,32{]} \\ \hline
2                & ConvGRU2D     & {[}5,5{]}     & 32               & None    & {[}H/2,W/2,32{]} & {[}H/2,W/2,32{]} \\ \hline
\multirow{4}{*}{3} & GAP        & {[}H/2,W/2{]} & -                & None    & {[}H/2,W/2,32{]} & {[}1,1,32{]}     \\ \cline{2-7} 
                   & Dense      & -             & 512              & None    & {[}32{]}         & {[}512{]}        \\ \cline{2-7} 
                   & Dense      & -             & 2                & None    & {[}512{]}        & {[}2{]}          \\ \cline{2-7} 
                   & Softmax    & -             & 2                & None    & {[}2{]}          & {[}2{]}          \\ \hline
\end{tabular}

\begin{tabular}{|l|l|l|l|l|l|l|}
\hline
\multicolumn{7}{|l|}{\textbf{CORNet-S}}                                                                                        \\ \hline
Layer              & Layer type & Kernel size   & N\_out & Pooling & Input            & Output           \\ \hline
1                  & Conv2D     & {[}7,7{]}     & 32               & Max     & {[}H,W,3{]}  & {[}H/2,W/2,32{]} \\ \hline
2                & CORBlock-S     & {[}5,5{]}     & 32               & None    & {[}H/2,W/2,32{]} & {[}H/2,W/2,32{]} \\ \hline
\multirow{4}{*}{3} & GAP        & {[}H/2,W/2{]} & -                & None    & {[}H/2,W/2,32{]} & {[}1,1,32{]}     \\ \cline{2-7} 
                   & Dense      & -             & 512              & None    & {[}32{]}         & {[}512{]}        \\ \cline{2-7} 
                   & Dense      & -             & 2                & None    & {[}512{]}        & {[}2{]}          \\ \cline{2-7} 
                   & Softmax    & -             & 2                & None    & {[}2{]}          & {[}2{]}          \\ \hline
\end{tabular}

\begin{tabular}{|l|l|l|l|l|l|l|}
\hline
\multicolumn{7}{|l|}{\textbf{V1Net-1L}}                                                                                        \\ \hline
Layer              & Layer type & Kernel size   & N\_out & Pooling & Input            & Output           \\ \hline
1                  & Conv2D     & {[}7,7{]}     & 32               & Max     & {[}H,W,3{]}  & {[}H/2,W/2,32{]} \\ \hline
2                & V1Net     & {[}5,5{]}     & 32               & None    & {[}H/2,W/2,32{]} & {[}H/2,W/2,32{]} \\ \hline
\multirow{4}{*}{3} & GAP        & {[}H/2,W/2{]} & -                & None    & {[}H/2,W/2,32{]} & {[}1,1,32{]}     \\ \cline{2-7} 
                   & Dense      & -             & 512              & None    & {[}32{]}         & {[}512{]}        \\ \cline{2-7} 
                   & Dense      & -             & 2                & None    & {[}512{]}        & {[}2{]}          \\ \cline{2-7} 
                   & Softmax    & -             & 2                & None    & {[}2{]}          & {[}2{]}          \\ \hline
\end{tabular}
\label{tab:rcnn_architectures}
\end{table} 